\documentclass[journal]{IEEEtran}

\usepackage{url}  
\usepackage{graphicx}  
\usepackage{amsmath,amssymb}
\usepackage{color}
\usepackage{epstopdf}
\usepackage{array}
\usepackage{multirow}
\usepackage{bbm}
\usepackage{amsfonts}
\usepackage{booktabs}
\usepackage{bm}
\usepackage{float}
\usepackage{subfigure}
\usepackage{cite}
\usepackage{ifpdf}
\usepackage{booktabs}
\usepackage[ruled, lined, longend, linesnumbered]{algorithm2e}

\usepackage{bbding}

\usepackage[breaklinks=true,colorlinks,bookmarks=true]{hyperref}

\def\rankS#1{\textit{{\color{blue}{#1}}}}

\begin{document}

\title{Dual-View Data Hallucination with Semantic Relation Guidance for Few-Shot Image Recognition}

\author{Hefeng Wu, Guangzhi Ye, Ziyang Zhou, Ling Tian, Qing Wang, Liang Lin,~\IEEEmembership{Fellow,~IEEE}
\thanks{This work was supported in part by National Natural Science Foundation of China (NSFC) under Grant No. 62272494, and in part by Guangdong Basic and Applied Basic Research Foundation under Grant No. 2023A1515012845 and 2023A1515011374. (Corresponding author: Liang Lin)}
\thanks{Hefeng Wu, Guangzhi Ye, Ziyang Zhou, Qing Wang, and Liang Lin are with the School of Computer Science and Engineering, Sun Yat-sen University, Guangzhou, China (e-mail: wuhefeng@gmail.com, yegzh@mail2.sysu.edu.cn, zhouzy36@mail2.sysu.edu.cn, ericwangqing@gmail.com, linliang@ieee.org).} 
\thanks{Ling Tian is with University of Electronic Science and Technology of China, Chengdu, China (e-mail: lingtian@uestc.edu.cn).}
}

\markboth{IEEE Transactions on Multimedia}%
{Wu \MakeLowercase{\textit{et al.}}: }

\maketitle

\begin{abstract}

Learning to recognize novel concepts from just a few image samples is very challenging as the learned model is easily overfitted on the few data and results in poor generalizability. One promising but underexplored solution is to compensate the novel classes by generating plausible samples. However, most existing works of this line exploit visual information only, rendering the generated data easy to be distracted by some challenging factors contained in the few available samples. Being aware of the semantic information in the textual modality that reflects human concepts, this work proposes a novel framework that exploits semantic relations to guide dual-view data hallucination for few-shot image recognition. The proposed framework enables generating more diverse and reasonable data samples for novel classes through effective information transfer from base classes. Specifically, an instance-view data hallucination module hallucinates each sample of a novel class to generate new data by employing local semantic correlated attention and global semantic feature fusion derived from base classes. Meanwhile, a prototype-view data hallucination module exploits semantic-aware measure to estimate the prototype of a novel class and the associated distribution from the few samples, which thereby harvests the prototype as a more stable sample and enables resampling a large number of samples. We conduct extensive experiments and comparisons with state-of-the-art methods on several popular few-shot benchmarks to verify the effectiveness of the proposed framework. 

\end{abstract}


\IEEEpeerreviewmaketitle

\section{Introduction}

\IEEEPARstart{T}{he} excellent performance of deep learning models \cite{GuoZJLL23tmm,SunSWJL22tmm,chen2022cross,CuiWZSW23tip,He2016ResNet} generally relies on a large amount of labelled data, which can be costly or infeasible to collect in diverse real-world applications. In contrast, human is able to learn new concepts from very few data samples. Motivated by this remarkable capacity of the human recognition system, few-shot learning (FSL) \cite{WangWJL2023CSVT,YangHLHW23tmm,WuCLCWL23ArXiv,ChenLCHW22pami,ZhangHZ21acmmm,LZLF2020CVPRAdversarial} has emerged as a popular research trend in recent years, which seeks to learn new concepts with minimal labeled data.

Intuitively, the main challenge of few-shot learning is the severe lack of training data, which easily causes the model to overfit on limited samples and exhibit poor generalization performance when encountering new test samples. 
This challenge is particularly pronounced when the available few-shot samples present visual complexities, such as low illumination or cluttered backgrounds with distracting objects.
Therefore, learning the common characteristics of a class is rather difficult in the context of few-shot learning.
Most existing FSL works gerenally follow the meta-learning paradigms \cite{vinyals_matching_2017,finnModelagnosticMetalearningFast2017,SungYZXTH18cvpr,LaiKHSS21tnn,chenNewMetaBaselineFewShot2020}, by designing a meta-learner of metrics or optimization schemes and so on to quickly adapt models to the new few-shot tasks.
Despite achieving much progress, these researches are still hard to obtain satisfying results.

Therefore, in this work we are dedicated to another direction with less attention but great potential, i.e., to compensate for the limited data by generation.
To make the model more robust and generalizable, this line of works \cite{ZhangCGBS18nips,SchwartzKSHMKFG18nips,yang2021free} design algorithms to generate new training data for few-shot classes,
based on the observations that certain patterns, such as intra-class deformations and distribution statistics, can generalize across classes.
By transferring such patterns from the base classes with abundant labeled samples, these algorithms can potentially create reasonable samples for data-starved classes. 
However, most of these methods only rely on visual information, which easily cause the generated samples to badly deviate from the true distribution and deteriorate the model, as the few-shot samples participating in the process may bring strong distracting bias.

\begin{figure}[!t]
    \centering
    \includegraphics[width=\linewidth]{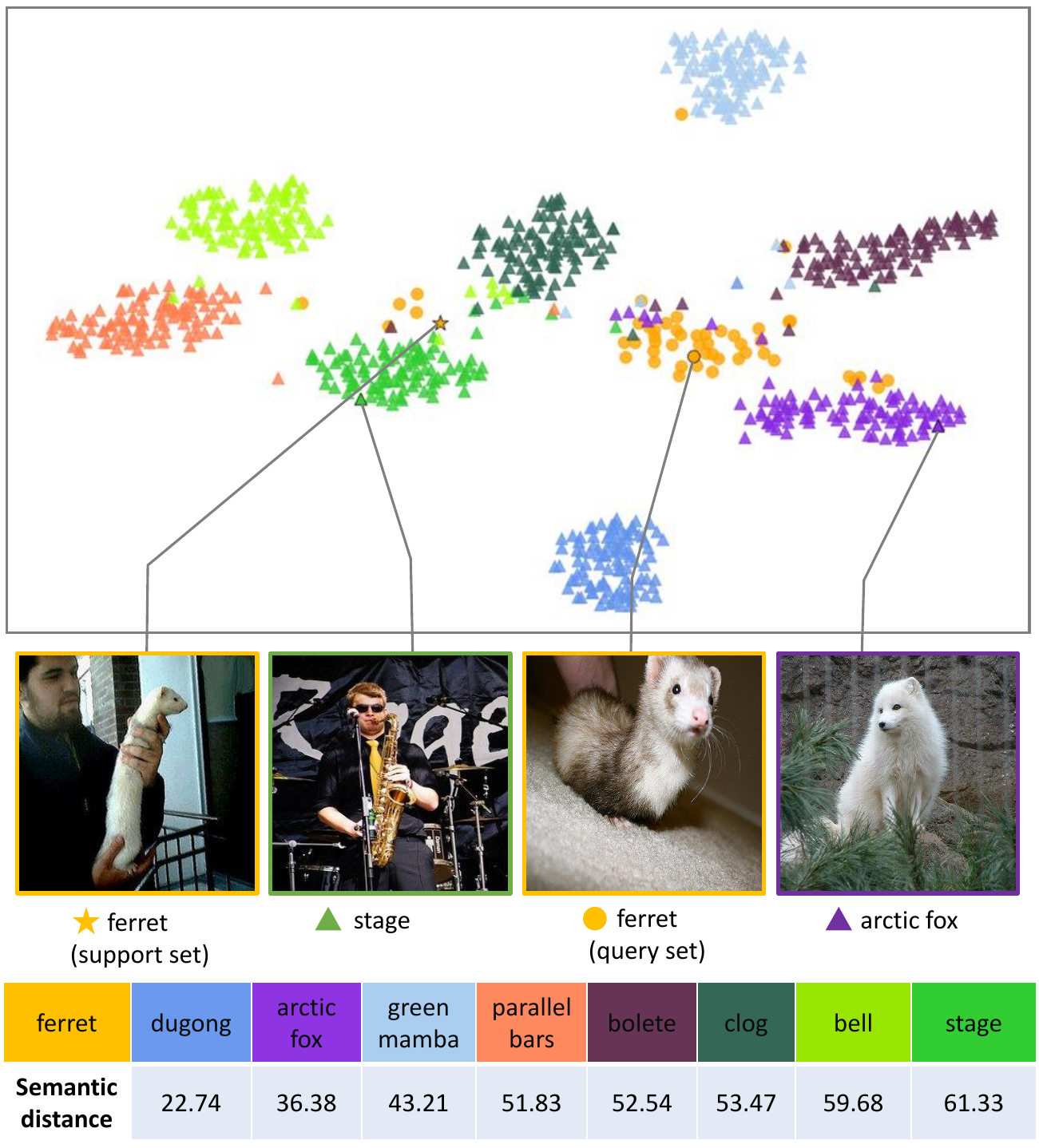}
    \caption{t-SNE visualization of image samples from several classes and their semantic distance with the ``ferret'' class. Each class is denoted with a different color and class ``ferret'' is assumed to have one-shot sample (support set). The ``arctic fox'' class is much closer to the ``ferret'' class in data distribution than the ``stage'' class, but it is not correctly reflected by the support sample of the ``ferret'' class in the visual space. In contrast, such relationships can be revealed from the semantic space. Best viewed in color.}
    \label{fig:motivation}
\end{figure}

Nevertheless, though only a few visual samples is available for a novel class, there exists useful semantic information in the  textual modality, which reflects human conceptual space and can provide extra knowledge from a complementary perspective. Such semantic information is contained in the given class label or additional descriptions that are easily acquired from lexical databases like WordNet \cite{1990WordNet}. 
Figure \ref{fig:motivation} illustrates an example of visual and semantic relationships between classes. Image samples from several classes are visualizd by the t-SNE algorithm \cite{Laurens2008Visualizing} to show their relationship in the visual feature space, with each class denoted by a different color.
The ``ferret'' class with yellow color is assumed to be a novel class with one-shot support sample (denoted with a star), and the other samples act as the query set.
Besides the support sample, three other image samples with different visual distances from it are also exhibited in Figure \ref{fig:motivation}. 
It can be observed that the support sample is distributed close to the cluster of the ``stage'' class, mainly because a man co-exists in the sample context, making it far away from the cluster center of the ``ferret'' class. In contrast, the ``arctic fox'' class is actually much closer to the ``ferret'' class in data distribution than the ``stage'' class, but it is not correctly reflected due to severely lacking data of the ``ferret'' class. 
Fortunately, such relationships can be revealed from the semantic perspective, as illustrated at the bottom of Figure \ref{fig:motivation}.
Motivated by these observations, therefore, we study incorporating semantic relations to guide the generation of new visual data for few-shot learning, giving birth to a novel framework, i.e., Dual-View Data Hallucination with Semantic Relation Guidance.

Specifically, the proposed framework transfers the information of base classes to generate diverse data samples of novel classes from both instance and prototype views. An instance-view data hallucination module uses semantic correlated attention maps to hallucinate each sample of a novel class locally in the spatial space, and also introduces global semantic feature fusion to generate new samples with higher confidence by projecting samples closer to the class prototype. 
However, although the instance-view module can generate some diverse samples for a novel class, the generated samples are not stable enough to represent the class and the sample number is still limited. Hence we further propose a prototype-view data hallucination module, which exploits semantic-aware measure to estimate the  prototypes of the novel classes as more stable new samples and also estimate the associated distributions for generating a theoretically unlimited number of data samples.

The contributions of this paper are summarized as follows.
First, we propose a novel dual-view data hallucination with semantic relation guidance framework for few-shot image recognition. Our proposed framework effectively exploits semantic relations to guide the generation of hallucinated data from both instance and prototype views. Its data-oriented paradigm is orthogonal to many existing FSL methods and can feasibly combine with them to yield better models.
Second, we design an instance-view data hallucination module to hallucinate each sample with local semantic correlated attention and global semantic feature fusion.
Third, we propose a prototype-view data hallucination module that estimates the prototypes of novel classes and the associated distributions under the guidance of semantic-aware measurement, which thereby harvests prototypes as more stable samples and enables resampling a sufficient number of samples. 
Moreover, extensive experiments and comparisons are conducted on several popular FSL datasets to demonstrate the superiority of the proposed framework and provide in-depth analysis.

The remainder of this paper is organized as follows. Section \ref{sec:RelatedWorks} reviews the related works, and Section \ref{sec:Methodology} introduces our proposed framework in detail. The experimental results are presented in Section \ref{sec:Experiments}, and finally Section \ref{sec:Conclusion} concludes the paper.

\section{Related works}\label{sec:RelatedWorks}

In this section, we review the related works from two perspectives, i.e., few-shot learning and learning with semantics.

\subsubsection{Few-shot learning}
Many existing FSL approaches follow the meta-learning paradigms, i.e., to learn a meta-learner from a lot of meta tasks that can quickly adapt to new few-shot tasks. In this work, we introduce the FSL works from a more concrete taxonomy and categorize them by four main perspectives: metric, optimization, model, and data. It is not a strict division since some works involve several perspectives.
(i) \textit{Metric-based} methods \cite{vinyals_matching_2017,SungYZXTH18cvpr,ZhangCLS23pamiDeepEMD,snellPrototypicalNetworksFewshot2017,YeHZS20cvpr,Afrasiyabi2022SetFeat,WangWJL2023CSVT,ZhouHHWGK2023TNNLS,ChengHHLZ23tmm} mainly focus on learning an embedding model that maps the original data into a representation space suitable for few-shot recognition, where same-class samples are close while different-class samples are faraway under a specific metric.
The Prototypical Network \cite{snellPrototypicalNetworksFewshot2017} is a classical method of this kind. It simply averages the few-shot embedded features of a novel class as the class prototype and recognizes a query sample by nearest-neighbor classifiers.
The cosine metric is combined with deep neural networks for few-shot learning in the Matching Net \cite{vinyals_matching_2017}, while the Mahalanobis distance is exploited in \cite{DasL20tip}. 
Zhang et al. \cite{ZhangCLS23pamiDeepEMD} explore differentiable Earth Mover's Distance as the metric function on top of deep networks.
(ii) \textit{Optimization-based} methods \cite{chenNewMetaBaselineFewShot2020,finnModelagnosticMetalearningFast2017,MunkhdalaiY17icml,lee2019meta,liuNegativeMarginMatters2020,RusuRSVPOH19iclr,mangla2020charting,OhYKY21iclr,BouniotRALH22eccv,ZhangLYF23pami,feiMELRMETALEARNINGMODELING2021} aim to develop optimization algorithms that can train a good model quickly with few-shot samples. 
Finn et al. \cite{finnModelagnosticMetalearningFast2017} propose a model-agnostic meta-learning algorithm to learn a good initialization for fast adaptation of deep networks in few-shot scenarios.
Rusu et al. \cite{RusuRSVPOH19iclr} propose to learn a low-dimensional latent embedding of model parameters and perform optimization-based meta-learning in this space.
Neg-Cosine \cite{liuNegativeMarginMatters2020} uses negative margin loss to learn more robust backbone. 
In \cite{feiMELRMETALEARNINGMODELING2021}, Fei et al. utilize cross-episode optimization constraint between meta-learning episodes with the same set of classes.
(iii) \textit{Model-based} methods \cite{Act2Param18cvpr,GidarisK19cvpr,LaiKHSS21tnn,liuEnsembleEpochWiseEmpirical2020a,AfrasiyabiLG21iccv,Yu2022AAAIHybrid} address the few-shot learning task by utilizing tailored models to generate the classifier parameters or designing sophisticated classification models. Lai et al. \cite{LaiKHSS21tnn} learn a task-adaptive classifier-predictor to generate classifier weights by a meta-learner. Qiao et al. \cite{Act2Param18cvpr} propose to directly predict parameters from the activation statistics. In \cite{GidarisK19cvpr}, a graph neural network denoising autoencoder model is developed to generate classification weights for few-shot learning. 
Liu et al. \cite{liuEnsembleEpochWiseEmpirical2020a} propose to meta-learn an ensemble of epoch-wise empirical Bayes models for robust predictions in few-shot tasks.
(iv) \textit{Data-based} methods \cite{LZLF2020CVPRAdversarial,HariharanG17iccv,XingROP19nips,ChenFZJXS19tip,ZhangZK19cvpr} overcome the lack of data by generating proper data for the novel classes directly.
Hariharan and Girshick \cite{HariharanG17iccv}  learn in-class transformations from base classes and transfer them to novel classes by a multi-layer perceptron model.
A delta-encoder is developed in \cite{SchwartzKSHMKFG18nips} to extract transferable intra-class deformations, which are then applied to novel classes for sample synthesis.
Yang et al. \cite{yang2021free} uses statistics of base classes to calibrate the data distribution of novel classes and generate new data. 
Zhang et al. \cite{ZhangZK19cvpr} utilize a saliency network to obtain the foregrounds and backgrounds of image samples and hallucinate data from foreground-background combinations. 
Li et al. \cite{LZLF2020CVPRAdversarial} propose an adversarial feature hallucination network that is built on a conditional Wasserstein Generative Adversarial Network (GAN) and takes the few labeled samples and random noise as input to hallucinate features. 
While most of these data-based methods solely exploit visual information, our work studies incorporating semantic relations to guide data generation from two views.

\subsubsection{Learning with semantics}

Though deep learning models are capable of learning strong representations from raw visual data, they lack the prior commonsense knowledge of humans that plays a critical role in the human visual recognition system. 
Therefore, recent researches \cite{YinZLWSZ22tmm,TangWWGDGC18pami,ChenXHWL19iccv,XieWSLWZ23tmm,WeiYang2019ICLR} explore textual semantics, a common form of human knowledge, to facilitate visual models and achieve impressive progress in various tasks.
Tang et al. \cite{TangWWGDGC18pami} improve semi-supervised object detection models by exploring object similarity knowledge from both visual and semantic domains.
Yang et al. \cite{WeiYang2019ICLR} address the visual navigation task by incorporating semantic priors into the deep reinforcement learning framework.
In \cite{ChenXHWL19iccv}, class-specific presentations are learned via label semantic guidance for multi-label image recognition, while pre-trained semantic embeddings are utilized in \cite{YinZLWSZ22tmm} to promote worldwide GPS coordinates.
In \cite{XieWSLWZ23tmm}, the label semantic information is utilized to generate self-adaptive hash centers for image retrieval.
Wang et al. \cite{WangYG18cvprSemantic} combine semantic embeddings and knowledge graphs in the graph convolutional network for zero-shot recognition.
Xian et al. \cite{XianLSA18cvpr} exploit a Wasserstein GAN conditioned on class-level fine-grained semantic information to synthesize CNN features for zero-shot learning.

There are also some works exploiting semantic information for few-shot image recognition \cite{LinWLC19ICIP,Yang0022wacv,ChenLLC22Shaping,HuangZKW21aaai}.
Yang et al. \cite{Yang0022wacv} propose a semantic guided attention mechanism for few-shot learning, where the semantic knowledge is used to guide visual perception to enhance the visual prototypes. 
Chen et al. \cite{ChenLLC22Shaping} present an attribute-shaped learning framework that predicts the attributes of queries and utilizes an attribute-visual attention module to produce more discriminative features for few-shot recognition.
In \cite{HuangZKW21aaai}, an attributes-guided attention module is devised to align pure-visual attention for fine-grained few-shot recognition. 
Inspired by these previous works, we propose a novel perspective on leveraging semantic information in few-shot learning. Our method explicitly harnesses the semantic relations as guidance to generate hallucinated data of novel classes from both instance and prototype views.
The work \cite{LinWLC19ICIP} also leverages semantic information to facilitate data generation for few-shot recognition, which bears resemblance to the GAN-based method \cite{XianLSA18cvpr} for zero-shot learning. Specifically, it employs a conditional GAN to generate data for few-shot classes, where semantic word vectors are utilized to generate the noise input for the GAN. 
However, our proposed method differs significantly from this work. Unlike \cite{LinWLC19ICIP}, our method does not involve the use of GANs for data generation. Instead, we use semantic relations to guide the data generation in a more efficient way through weighting and sampling.

\section{Methodology}\label{sec:Methodology}

\begin{figure*}
\centering
  \includegraphics[width=0.99\textwidth]{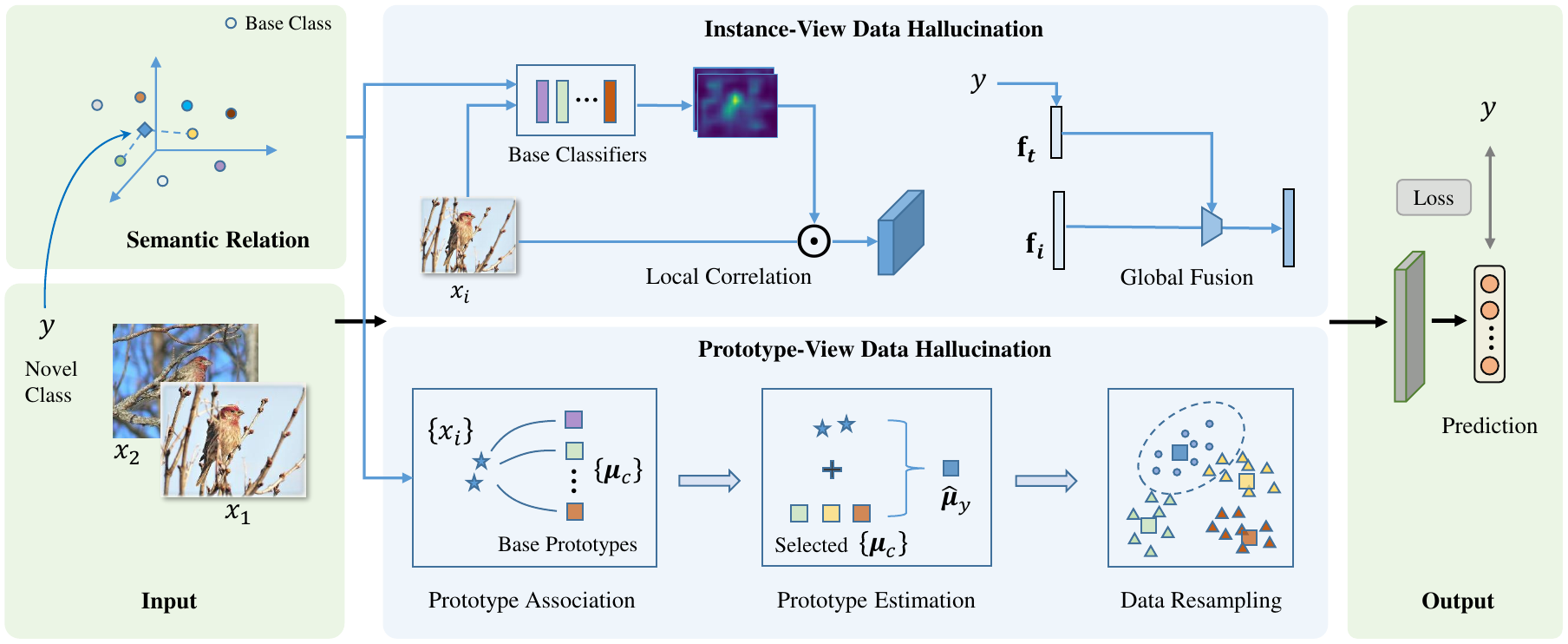}
  \caption{Illustration of our data hallucination framework. It generates hallucinated data of novel classes from both instance and prototype views with semantic relation guidance to facilitate model training. The instance-view data hallucination module generates new samples from each instance sample by local semantic correlation and global semantic fusion. The prototype-view data hallucination module explores semantic-aware measure to estimate the prototype of a novel class and the associated distribution, which thereby harvests the prototype as a more stable sample and enables resampling a sufficient number of samples. Finally, the hallucinated data are combined with the original data to train the recognition model. }
  \label{fig:framework}
\end{figure*}

\subsection{Problem Definition}
We follow previous work \cite{vinyals_matching_2017,snellPrototypicalNetworksFewshot2017,ZhangCLS23pamiDeepEMD} to formulate the few-shot image recognition task. Given a dataset $\mathcal{D} = \{(x_i, y_i)\}$ where $x_i$ denotes an image sample and $y_i \in \mathcal{C}$ is its  class label. $\mathcal{C}$ denotes the set of class labels. In the few-shot setting, $\mathcal{D}$ consists of two parts, i.e., $\mathcal{D}=\mathcal{D}_{base} \cup \mathcal{D}_{novel}$, where $\mathcal{D}_{base}$ and $\mathcal{D}_{novel}$ are the datasets of base and novel classes, respectively. Their corresponding label sets $\mathcal{C}_{base}$ and $\mathcal{C}_{novel}$ are disjoint, i.e., $\mathcal{C}_{base} \cup \mathcal{C}_{novel} = \mathcal{C}$ and $\mathcal{C}_{base} \cap \mathcal{C}_{novel} = \emptyset$. A few-shot image recognition task $\mathcal{T}=(\mathcal{S}, \mathcal{Q})$ is formulated in a $N$-way $K$-shot form, where the support set $\mathcal{S} = \{(x_i,y_i)\}_{i=1}^{N\times K}$ contains $N$ novel classes from $\mathcal{D}_{novel}$ with $K$ image samples in each class and the query set $\mathcal{Q} = \{(x_i,y_i)\}_{i=1}^{N\times M}$ contains $N\times M$ test samples of these classes. The goal is to learn a good classifier on the few-shot support set $\mathcal{S}$, along with the auxiliary base dataset $\mathcal{D}_{base}$ that has abundant labeled samples, so that the learned model can work well on the query set $\mathcal{Q}$. 
In evaluation, in order to comprehensively estimate the performance of a few-shot learning approach, $\mathcal{T}$ is sampled many times from $\mathcal{D}_{novel}$ and the approach's performance is measured by the average accuracy on these few-shot tasks.

\subsection{Framework Overview}
In this work, a simple model is adopted for image recognition, which consists of a backbone network for feature extraction and a classifier for prediction. 
In prevailing learning paradigms, obtaining a proficient representation of a given visual concept generally requires a substantial number of samples. In scenarios where image samples are scarce, this presents a considerable challenge. 
The proposed dual-view data hallucination framework aims to exploit semantic relation guidance to complement the few-shot samples with hallucinated data to learn a good recognition model.
Figure \ref{fig:framework} illustrates our proposed framework, which consists of two main modules, i.e., instance-view and prototype-view data hallucination modules. Specifically, the instance-view data hallucination module generates hallucinated samples from each instance sample by employing local semantic correlation and global semantic fusion, while the prototype-view data hallucination module exploits semantic-aware  measure to guide estimating the prototype of a novel class and the associated distribution, providing the prototype as a more stable sample and enabling resampling a sufficient number of samples. 
Finally, the hallucinated data are combined with the support set to train the recognition model.

\subsection{Feature Representation and Class Prototype}
\label{sec:Representation}

\subsubsection{Visual feature}
The backbone network of the recognition model can be viewed as a feature extractor of visual information, denoted as $\phi$. Given an image sample $x_i$, the feature extractor obtains its feature $\mathbf{f}_i \in \mathbf{R}^d$, formulated as
\begin{equation}
    \mathbf{f}_i = \phi(x_i).
\end{equation}
For learning good feature representation, the feature extractor is generally pre-trained on the auxiliary base dataset $\mathcal{D}_{base}$ that has sufficient training samples.

\subsubsection{Semantic representation}
Our proposed framework uses semantic information to guide generating hallucinated samples for training and involves semantic feature extraction from texts. 
Let $\phi_{sr}$ denote the semantic feature extractor.
For each class $c$, we extract a semantic class vector $\mathbf{v}_c\in \mathbf{R}^m$ using the pre-trained semantic feature extraction model $\phi_{sr}$, formulated as
\begin{equation}
     \mathbf{v}_c = \phi_{sr}(S_c),
\end{equation}
where $S_c$ is the semantic label text for class $c$.

\subsubsection{Class prototype}
We follow previous work \cite{snellPrototypicalNetworksFewshot2017} to use the mean of feature vectors of all samples from a class $c$ as the class prototype when the number of samples is sufficient. The prototype $\boldsymbol{\mu}_c$ of class $c$ is calculated as follows
\begin{equation}
    \boldsymbol{\mu}_c = \frac{\sum_{j=1}^{n_c}\mathbf{f}_j}{n_c},
\end{equation}
where $\mathbf{f}_j$ is the feature of the $j$th image sample of class $c$, and $n_c$ is the sample number.

We assume the data distribution of a class is Gaussian for simplicity.
With this assumption, the covariance matrix $\boldsymbol{\Sigma}_c$ of class $c$ is estimated as
\begin{equation}
    \boldsymbol{\Sigma}_c = \frac{\sum_{j=1}^{n_c}(\mathbf{f}_j-\boldsymbol{\mu}_c)(\mathbf{f}_j-\boldsymbol{\mu}_c)^T}{n_c-1}.
\end{equation}

\subsection{Instance-View Data Hallucination}\label{sec:IVDH}
The instance-view data hallucination (IVDH) module hallucinates each sample of a novel class to generate new data from local and global aspects with semantic relation guidance.

\subsubsection{Local semantic correlation}\label{sec:SCA}
As mentioned previously, semantically correlated concepts generally share similar visual appearance, which stems from human cognitive system. We propose to hallucinate an instance sample locally by using the spatial attention maps activated from the classifiers of base classes that are highly correlated with the corresponding novel class in semantics. Specifically, we generate the attention maps by Grad-CAM \cite{SelvarajuCDVPB17iccv} and then use them to hallucinate the samples. The Grad-CAM method is widely used in computer vision for model interpretability analysis. It can obtain the activation map that reflects the important regions used by the model to predict the class label of the input image.

Given a sample $(x,y)$ from a novel class, we can compute its semantic distance with a base class $c$ as follows
\begin{equation}\label{eq:semanticDis}
    d_s(y,c) = \| \mathbf{v}_y -\mathbf{v}_{c} \|^2,\;  c \in \mathcal{C}_{base} 
\end{equation}
where $\mathbf{v}_y$ and $\mathbf{v}_{c}$ denote the semantic features of classes $y$ and $c$, respectively.

Let $\mathrm{top}_{k}(i\in\mathcal{S}\,|\,\Phi(i))$ denote the top $k$ elements in the set $\mathcal{S}$ under the metric $\Phi$ (largest ranks first).
Then the top $k$ base classes most similar with the novel class $y$ in semantics is given by
\begin{equation}\label{eq:topkSet}
   \mathcal{S}_y =  \mathrm{top}_{k}(c \in \mathcal{C}_{base}| -d_s(c, y)).
\end{equation}

When pre-training the backbone network on the base dataset $\mathcal{D}_{base}$, we also obtain the classifiers for the base classes. By inputing the sample $x$ to the model, we can compute the activation maps from base classifiers by Grad-CAM.

For a base class $c\in\mathcal{S}_y$, let $M_x^c$ denote its computed activation map for sample $x$. Then a hallucinated image sample $x'$ is generated as follows
\begin{equation}
    x' = ((M_x^c)^t+1) \odot x,
\end{equation}
where $\odot$ denotes element-wise multiplication, and $(\cdot)^t$ denotes element-wise exponentiation with $t$ that aims to smooth the attention map. In our implementation $t$ is set to 0.5. 
The new sample $x'$ is then further processed by a commonly used random augmentation operation, including random size crop, image jitter, random flip, etc.
The local semantic correlated attention can emphasize the image parts activated by semantically similar base classes, providing hallucinated samples to lead the model to focus on useful features for recognizing the novel classes. However, less similar base classes may also bring noise, so in experiments, we empirically set $k$=1 to retrieve the most  similar base class.

\subsubsection{Global semantic fusion}\label{sec:SFI}

We propose a simple global semantic feature fusion network that learns to project a sample closer to the prototype of its class in the feature space with semantic embedding fusion, thereby generating  hallucinated feature samples with higher confidence for novel classes.

The fusion network is learned on the base dataset $\mathcal{D}_{base}$.
Given a training image sample $x_i$ of class $y_i\in\mathcal{C}_{base}$, we extract its visual feature $\mathbf{f}_i$ and the semantic feature $\mathbf{v}_{y_i}$, as described in Section \ref{sec:Representation}.
We introduce a semantic fusion mechanism to guide learning a feature vector that is closer to its class prototype for $x_i$. Specifically, we first learn a hidden feature vector $\mathbf{h}_i$ by mapping from both the visual feature $\mathbf{f}_i$ and the semantic feature $\mathbf{v}_{y_i}$:
\begin{equation}
    \mathbf{h}_i = \varphi(\mathbf{f}_i \oplus \mathbf{v}_{y_i}),
\end{equation}
where $\oplus$ denoted the concatenation operation and  $\varphi$ is a simple mapping function that is implemented by a fully connected layer.

Next the hidden feature $\mathbf{h}_i$ is merged with the original feature $\mathbf{f}_i$ in a weighted residual connection scheme to obtain the final semantically fused feature $\mathbf{\hat{f}}_i$, formulated as
\begin{equation}\label{eq:fuse}
    \mathbf{\hat{f}}_i = \text{ReLU}(\mathbf{f}_i + \lambda \cdot \sigma(\mathbf{h}_i)),
\end{equation}
where $\sigma$ denoted the tangent activation function, $\text{ReLU}$ is the rectified linear unit function, and $\lambda$ is a hyper-parameter that controls the strength of semantic embedding fusion.

We aim to push $\mathbf{\hat{f}}_i$ towards the prototype feature $\boldsymbol{\mu}_{y_i}$ of class ${y_i}$, and the following mean square error (MSE) loss is adopted as the objective function for learning the fusion network
\begin{equation}
    \mathcal{L}_s = \frac{1}{n}\sum_{i=1}^{n}\|\mathbf{\hat{f}}_i - \boldsymbol{\mu}_{y_i} \|^2,
\end{equation}
where $n$ is the number of training samples and $y_i$ is the class label of the $i$th sample.

\begin{table*}[!t]
\centering
\caption{Few-shot classification accuracy(\%) on \textit{mini}ImageNet, \textit{tiered}ImageNet and CUB with 95\% confidence intervals. The best and second best results are highlighted in bold and italic blue texts, respectively. A, B, C and D represent the corresponding methods as metric-, optimization-, model-, and data-based, respectively. ``-'' denotes the corresponding result is not provided. }\label{tab:sota}
\begin{tabular}{c|c|c|c|cc|cc|cc}
\toprule
\multirow{2}{*}{Method} & \multirow{2}{*}{Year} & \multirow{2}{*}{Type} & \multirow{2}{*}{Backbone} & \multicolumn{2}{c|}{\textit{mini}ImageNet} & \multicolumn{2}{c|}{\textit{tiered}ImageNet} & \multicolumn{2}{c}{CUB}  \\
                         &   &  &                         & 5-way  1-shot     & 5-way 5-shot               & 5-way 1-shot       & 5-way 5-shot               & 5-way 1-shot     & 5-way 5-shot      \\
\midrule
Act2Param\cite{Act2Param18cvpr}  & 2018  & C & WRN-28-10                & 59.60 ± 0.41 & 73.74 ± 0.19           & -            & -                    & -          & -           \\
LEO\cite{RusuRSVPOH19iclr}                       & 2018  & B & WRN-28-10                & 61.76 ± 0.08 & 77.59 ± 0.12           & 66.33 ± 0.05   & 81.44 ± 0.09           & 68.22 ± 0.22 & 78.27 ± 0.16  \\
DualTriNet\cite{ChenFZJXS19tip}                      & 2019  & D & ResNet18               & 58.12 ± 1.37 & 76.92 ± 0.69           & -   & -           & 69.61 ± 0.46       & 84.10 ± 0.35        \\
AM3\cite{XingROP19nips}                      & 2019 & D & ResNet12                & 65.30 ± 0.49 & 78.10 ± 0.36           & 69.08 ± 0.47   & 82.58 ± 0.31           & 74.1       & 79.7        \\
MetaOptNet \cite{lee2019meta} & 2019  & B & ResNet12               & 64.09 ± 0.62 & 80.00 ± 0.45           & 65.81 ± 0.74 & 81.75 ± 0.53         & -          & -           \\
GNN-DAE\cite{GidarisK19cvpr} & 2019  & C & WRN-28-10              & 61.07 ± 0.15 & 76.75 ± 0.11           & 68.18 ± 0.16   & 83.09 ± 0.12           & -          & -           \\
Neg-Cosine\cite{liuNegativeMarginMatters2020}               & 2020 & B & ResNet18                & 62.33 ± 0.82 & 80.94 ± 0.59           & -            & -                    & 72.66 ± 0.85 & 89.40 ± 0.43  \\
Meta-Baseline\cite{chenNewMetaBaselineFewShot2020}              & 2020 & B & ResNet12                 & 63.17 ± 0.23 & 79.26 ± 0.17           & 68.62 ± 0.27   & 83.29 ± 0.18           & -          & -           \\
E3BM\cite{liuEnsembleEpochWiseEmpirical2020a}                     & 2020  & C & ResNet18               & 64.70 ± 1.80 & 81.20 ± 0.60           & 70.70 ± 1.70   & 84.90 ± 0.80           & -          & -           \\
S2M2R\cite{mangla2020charting}                    & 2020   & B & WRN-28-10               & 64.93 ± 0.18 & 83.18 ± 0.11           & 73.71 ± 0.22   & \rankS{88.59 ± 0.14}           & 80.68 ± 0.81 & \rankS{90.85 ± 0.44}  \\
FEAT \cite{YeHZS20cvpr}                     & 2020  & A & ResNet12                & 66.78 ± 0.20 & 82.05 ± 0.14           & 70.80 ± 0.23   & 84.79 ± 0.16           & -          & -           \\
AFHN \cite{LZLF2020CVPRAdversarial}           & 2020  & D & ResNet18                & 62.38 ± 0.72 & 78.16 ± 0.56           & -          & -           & 70.53 ± 1.01   & 83.95 ± 0.63           \\
BOIL \cite{OhYKY21iclr}                      & 2021  & B & ResNet12              & 49.61 ± 0.16      & 66.45 ± 0.37                & 48.58 ± 0.27        & 69.37 ± 0.12                & 61.60 ± 0.57          & 75.96 ± 0.17           \\
MPM\cite{LaiKHSS21tnn}                      & 2021     & C & WRN-28-10           & 61.77      & 78.03                & 67.58        & 83.93                & -          & -           \\
MixtFSL \cite{AfrasiyabiLG21iccv}                      & 2021  & C & WRN-28-10              & 64.31 ± 0.79      & 81.66 ± 0.60                & 70.97 ± 1.03        & 86.16 ± 0.67                & 73.94 ± 1.10          & 86.01 ± 0.50            \\
MELR\cite{feiMELRMETALEARNINGMODELING2021}                       & 2021    & B & ResNet12              & 67.40 ± 0.43 & 83.40 ± 0.28           & 72.14 ± 0.51   & 87.01 ± 0.35           & -          & -           \\
DC\cite{yang2021free}         & 2021  & D & WRN-28-10                & 68.43 ± 0.63 & 84.01 ± 0.43           & \rankS{75.97 ± 0.67}   & 88.05 ± 0.47           & {79.75 ± 0.68} & 90.80 ± 0.35  \\
MTR\cite{BouniotRALH22eccv}         & 2022   & B & ResNet12               & 62.69 ± 0.20 & 80.95 ± 0.14           & 68.44 ± 0.23   & 84.20 ± 0.16           & - & -  \\
HGNN\cite{Yu2022AAAIHybrid}         & 2022   & C & ResNet12               & 67.02 ± 0.20 & 83.00 ± 0.13           & 72.05 ± 0.23   & 86.49 ± 0.15           & 78.58 ± 0.20 & 90.02 ± 0.12  \\
SetFeat\cite{Afrasiyabi2022SetFeat}         & 2022     & A & ResNet12             & 68.32 ± 0.62 & 82.71 ± 0.46           & 73.63 ± 0.88   & 87.59 ± 0.57           & 79.60 ± 0.80 & 90.48 ± 0.44  \\
Auto-MS\cite{ZhouHHWGK2023TNNLS}         & 2023     & A & HCE-64F             & 53.33 ± 0.81 & 69.64 ± 0.91           & 57.25 ± 0.91   & 77.46 ± 0.71           & - & -  \\
QSFormer\cite{WangWJL2023CSVT}         & 2023     & A & ResNet12             & 65.24 ± 0.28 & 79.96 ± 0.20           & 72.47 ± 0.31   & 85.43 ± 0.22           & 75.44 ± 0.29 & 86.30 ± 0.19  \\
PCWOPK \cite{ZhangLYF23pami}                   & 2023  & B & ResNet12                &  67.10 ± 0.84          & 81.35 ± 0.56 & 72.49 ± 0.95   & 85.46 ± 0.63           & \textbf{82.04 ± 0.70} & 90.58 ± 0.50  \\
DeepEMD \cite{ZhangCLS23pamiDeepEMD}                   & 2023  & A & ResNet12                &  \rankS{68.77 ± 0.29}          & \rankS{84.13 ± 0.53} & 74.29 ± 0.32   & 86.98 ± 0.60           & 79.27 ± 0.29 & 89.80 ± 0.51  \\
MBSS \cite{ChengHHLZ23tmm}                   & 2023  & A & WRN-28-10                & 66.91 ± 0.20 & 82.41 ± 0.14           & 72.81 ± 0.23   & 86.81 ± 0.15           & 75.49 ± 0.69 & 89.06 ± 0.40  \\ 
\hline
Ours                     & -  & D & WRN-28-10               & \textbf{70.64 ± 0.60} & \textbf{84.71 ± 0.41}           & \textbf{76.43 ± 0.67}   & \textbf{89.67 ± 0.44}           & \rankS{81.57 ± 0.61} & \textbf{91.30 ± 0.32}  \\
\toprule
\end{tabular}
\end{table*}

\subsection{Prototype-View Data Hallucination} \label{sec:PVDH}
The prototype-view data hallucination (PVDH) module exploits semantic-aware measure between the novel and base classes to guide estimating the prototypes of novel classes and the associated distributions, thereby harvesting the prototypes as more stable samples and enabling resampling a number of samples for model training. 
It is based on the observations that more correlated classes generally have more similar means and variances of sample features \cite{yang2021free}.

\subsubsection{Prototype association}\label{sec:PA} Specifically, given an image sample $x_i$ from a novel class $y$, we first design a semantic-visual similarity metric to find the most correlated base prototypes for it. 

Let $\mathbf{f}_i$ denote the visual feature of $x_i$, following \cite{yang2021free}, we apply Tukey’s Ladder of Powers transformation \cite{tukey1977exploratory} to $\mathbf{f}_i$, making its potential distribution more Gaussian-like, formulated as  
\begin{equation}
    \mathbf{f}'_i = \mathbf{f}_i^{\tau},
\end{equation}
where ${\tau}$ is a hyper-parameter, and $\mathbf{f}'_i=\log(\mathbf{f}_i)$ if ${\tau}=0$.

Following Equation (\ref{eq:semanticDis}), we compute the visual distance of sample $x_i$ with a base class $c$ as
\begin{equation}\label{eq:visualDis}
    d_v(x_i,c) = \| \mathbf{f}'_i -\boldsymbol{\mu}_c \|^2,\;  c \in \mathcal{C}_{base} 
\end{equation}
where $\boldsymbol{\mu}_c$ denotes the class prototype of $c$.

Different similarity metrics can be designed by combining the semantic distance $d_s(y,c)$ in Equation (\ref{eq:semanticDis}) and the visual distance $d_v(x_i,c)$  in Equation (\ref{eq:visualDis}). Without the need to pay special attention to the magnitude balance of the two distances, we propose a simple semantic-visual similarity metric via ranking to select a set of $q$ most correlated base classes $\mathcal{B}_{i}$, formulated as
\begin{equation}\label{eq:semanticVisualDis}
\begin{aligned}
   \mathcal{B}_{i} &=  \mathrm{top}_{q}(c \in \mathcal{T}_{y}| -d_v(x_i, c)), \\ 
   \mathcal{T}_{y} &=\mathrm{top}_{p}(c \in \mathcal{C}_{base}| -d_s(c, y)),
\end{aligned}
\end{equation}
where the $\mathrm{top}(\cdot)$ operator is defined in Equation (\ref{eq:topkSet}), and $p$ is a hyper-parameter larger than $q$ to control the scale of semantic selection.

\subsubsection{Prototype estimation}\label{sec:PE}
Afterwards, a candidate prototype of class $y$ is computed by fusing the class prototypes of the selected base classes $\mathcal{B}_{i}$ and the sample feature $\mathbf{f}'_i$, formulated as 
\begin{equation}
    \boldsymbol{\mu}'_i = \alpha \frac{\sum_{c \in \mathcal{B}_{i}} \boldsymbol{\mu}_c }{q} + (1-\alpha) \mathbf{f}'_i \,,
\end{equation}
where $\alpha$ is a parameter that controls the degree of prototype fusion. 

Given $K$ samples of the novel class $y$, we repeat the above procedure for every sample and obtain the corresponding candidate prototypes, i.e., $\{\boldsymbol{\mu}'_i\}_{i=1}^K$. 
We combine them by averaging to obtain the final estimation of the class prototype, formulated as
\begin{equation}
    \boldsymbol{\hat{\mu}}_y = \frac{1}{K}\sum_{i=1}^K\boldsymbol{\mu}'_i.
\end{equation}

\subsubsection{Data resampling}\label{sec:DR}
Similarly, we can compute its associated covariance matrix as follows 
\begin{equation}
     \boldsymbol{\hat{\Sigma}}_y = \frac{1}{K}\sum_{i=1}^K\boldsymbol{\Sigma}'_i \,, \quad 
    \boldsymbol{\Sigma}'_i = \frac{\sum_{c\in \mathcal{B}_{i}} \boldsymbol{\Sigma}_c}{q}+\beta \boldsymbol{1} \,, 
\end{equation}
where $\boldsymbol{1}$ is an all-one matrix with the same size as $\boldsymbol{\Sigma}$, and $\beta$ is a parameter that provides basic variance of features, which is set as 0.2 empirically.

The above estimations can benefit model training from two aspects.
On the one hand, the estimated prototypes of novel classes can serve as more stable hallucinated samples.
On the other hand, with the Gaussian distribution assumption, a large number of samples can be re-sampled from the estimated distributions, e.g., randomly generating $\mathbf{f} \sim  \mathcal{N}(\boldsymbol{\hat{\mu}}_y, \boldsymbol{\hat{\Sigma}}_y')$ for class $y$.

\subsection{Optimization Loss}

The recognition model follows a conventional multi-class classification paradigm. With a sample feature $\mathbf{f}_i$, the recognition model will output a predicted probability vector $\mathbf{p}_i=(p_{i1}, p_{i2}, \dots, p_{iN})$, where $N$ is the number of novel classes and $p_{ic}$ denotes the probability of the sample belonging to class $c$. The classifier is generally implemented as a fully-connected layer (i.e., linear classifier) followed by a softmax function.

We adopt the cross entropy as the objective loss function for training:
\begin{equation}
  \mathcal{L}=-\sum_{i=1}^{M}\sum_{c=1}^{N} y_{ic}\log p_{ic},
\end{equation}
where $M$ is the number of samples, including the original and hallucinated ones, and $(y_{i1}, y_{i2}, \dots, y_{iN})$ is a one-hot vector that $y_{ic}$ is 1 if the $i$th sample belongs to class $c$.

\section{Experiments}\label{sec:Experiments}

\subsection{Experimental Settings}\label{sec:settings}
\subsubsection{Datasets} 
We use three mainstream public FSL datasets in our experiments, i.e., \textit{mini}ImageNet \cite{vinyals_matching_2017}, \textit{tiered}ImageNet \cite{RenTRSSTLZ18iclr} and CUB \cite{2010Caltech} datasets, respectively.
 
{\textit{mini}ImageNet} \cite{vinyals_matching_2017} is a subset sampled from the ImageNet dataset \cite{2009ImageNet}. It consists of 100 classes with 600 image samples in each class. The size of images is $84\times 84$. The dataset is split into 64 base classes, 16 validation classes and 20 novel classes. 

{\textit{tiered}ImageNet} \cite{RenTRSSTLZ18iclr} is a larger subset of ImageNet compared with \textit{mini}ImageNet. It totally contains 608 classes, including 351 base classes, 97 validation classes and 160 novel classes. The average number of images in each class is 1281. In this dataset, the classes are selected with a hierarchical structure to ensure that base and novel classes do not belong to the same higher-level categories. Therefore, \textit{tiered}ImageNet ensures that base and novel classes are split semantically and share less similarity, which requires stronger robustness of the model.
 
{CUB} \cite{2010Caltech} is a fine-grained dataset comprising 200 kinds of birds, with 11,788 images in total. The size of images is $84\times 84$. The CUB dataset is split into 100 base classes, 50 validation classes and 50 novel classes.

\subsubsection{Evaluation Protocols}
Following most previous works, we conduct experiments in the 5-way 1-shot/5-shot settings for evaluation and comparison. Specifically, 1000 tasks are randomly sampled from the novel classes for testing and then the mean classification accuracy is computed, as well as the 95\% confidence interval.

\subsubsection{Implementation Details}
We adopt WRN-28-10 \cite{ZagoruykoK16bmvc} as the backbone (visual feature extractor), which is commonly used in few-shot learning tasks. The backbone ends with ReLU activation.
For each dataset, the backbone is pre-trained on the base classes. 
The output feature dimension is 640. The classifier is learned using the scikit-learn toolbox \cite{PedregosaVGMTG11jmlr}.
For the semantic feature extractor $\phi_{sr}$, we employ DistilBERT \cite{Sanh2019DistilBERTAD}, which is a distilled version of BERT \cite{devlin2018bert} with a reduced weight and faster computation speed while retaining comparable performance. To make better use of semantic information, we use the class label to get its definition in WordNet \cite{1990WordNet} and combine the two as the semantic text $S_c$ if the corresponding definition exists.
In the IVDH module, we utilize Grad-CAM \cite{SelvarajuCDVPB17iccv} to generate spatial attention maps for hallucinating the samples.  
We train the fusion network on the base classes for 100,000 iterations, using SGD optimizer with batch size of 5 and learning rate as 0.01. The parameter $\lambda$ is set as 0.3.
For the PVDH module, the parameter $\tau$ for Turkey's Ladder of Powers transformation is set as 0.5 by following previous works \cite{tukey1977exploratory,yang2021free}. We set $q$=2 to select similar base classes with $p$=10/32 for 1-shot/5-shot settings, and set $\alpha$=0.6  for prototype estimation. The number of resampled features is 200 per class.

\begin{table}
\centering
\caption{cross-domain evaluation from \text{mini}ImageNet to CUB.}\label{tab:cross_domain}
\begin{tabular}{c|c|c}
\toprule
Method                                        & 5-way 1-shot       & 5-way 5-shot       \\ \midrule
DCO\cite{lee2019meta}                     & 44.79  ±  0.75 & 64.98  ±  0.68 \\
FT\cite{tseng2020cross}                   & 47.47  ±  0.75 & 66.98  ±  0.68 \\
ATA\cite{WangD21ijcai}                   & \textbf{50.26  ±  0.50} & 65.31  ±  0.40 \\
Ours                             & 48.06  ±  0.62 & \textbf{70.01  ±  0.54} \\
\toprule
\end{tabular}
\end{table}

\subsection{Comparison to State-of-the-art Methods}

We compare our proposed method with state-of-the-art few-shot learning methods to demonstrate its effectiveness. 
Table \ref{tab:sota} reports the comparison results of our method and these FSL methods on the \textit{mini}ImageNet, \textit{tiered}ImageNet and CUB datasets, respectively. 
Following the compared methods, we provide evaluation results under 5-way 1-shot and 5-way 5-shot settings. 
As exhibited in Table \ref{tab:sota}, on the \textit{mini}ImageNet dataset that is most widely used for few-shot learning tasks, the previous best performing method DeepEMD achieves the classification accuracy of 68.77\% and 84.13\% for the 1-shot and 5-shot scenarios, respectively. Compared with DeepEMD, our method further improves the accuracy by 1.87\% and 0.58\% for the 1-shot/5-shot scenarios, leading to new state-of-the-art results. With respect to the larger and more challenging \textit{tiered}ImageNet dataset, our proposed method achieves the accuracy of 76.43\% and 89.67\% for the 1-shot and 5-shot settings, improving over the previous best performing methods by 0.46\% and 1.62\%, respectively. 
On the fine-grained CUB dataset, our method achieves the second best performance in the 1-shot setting, with the accuracy slightly lower than that of PCWOPK. While in the 5-shot setting, our method outperforms all the competitors, obtaining a performance gain of 0.45\% over the second best method S2M2R. 
It is noteworthy that, compared to other state-of-the-art data-based competitors, our method outperforms them by noticeable margins on all three datasets.
Overall, these comprehensive comparison results with state-of-the-art methods highlight the effectiveness of our method.

\subsection{Cross-Domain Evaluation}

In actual scenarios, there may be significant domain differences between base classes and novel classes. Therefore, we conduct experiments to analyze the performance of our method in cross-domain scenarios. 

First, we follow previous work \cite{lee2019meta,tseng2020cross} to conduct a cross-domain experiment from \text{mini}ImageNet to CUB. Namely, the base classes here comes from \text{mini}ImageNet, and the novel classes comes from CUB, which reflects the migration from coarse-grained classification to fine-grained classification. As shown in Table \ref{tab:cross_domain}, our method achieves good performance compared with leading cross-domain few-shot learning methods.

\begin{table}[!t]
\centering
\caption{Cross-domain evaluation on Meta-Dataset.} \label{tab:metadataset}
\resizebox{1.01\linewidth}{!}{
\begin{tabular}{c|c|c|c|c}
\toprule 
Method                    & MSCOCO          & Textures  & Fungi  & Omniglot                   \\ \midrule
k-NN \cite{triantafillou2019meta}                  & 30.38 ± 0.99        & 66.36 ± 0.75 & 36.16 ± 1.02     & 37.07 ± 1.15     \\
MatchingNet \cite{vinyals_matching_2017}            & 34.99 ± 1.00        & 64.15 ± 0.85 & 33.97 ± 1.00     & 52.27 ± 1.28          \\
ProtoNet \cite{snellPrototypicalNetworksFewshot2017}               & 41.00 ± 1.10        & 66.56 ± 0.83 & 39.71 ± 1.11     & 59.98 ± 1.35          \\
fo-MAML \cite{finnModelagnosticMetalearningFast2017}                & 35.30 ± 1.23        & 68.04 ± 0.81 & 32.10 ± 1.10     & 55.55 ± 1.54          \\
RelationNet \cite{SungYZXTH18cvpr}            & 29.15 ± 1.01        & 52.97 ± 0.69 & 30.55 ± 1.04     & 45.35 ± 1.36          \\
fo-Proto-MAML \cite{triantafillou2019meta}          & 43.74 ± 1.12        & 66.49 ± 0.83 & 39.96 ± 1.14     & 63.37 ± 1.33          \\
BOHB \cite{saikia2020optimized}                  & 48.03 ± 0.99        & 68.34 ± 0.76 & 41.38 ± 1.12     & \textbf{67.57 ± 1.21}          \\
SimpleCNAPS \cite{bateni2020improved}      & 45.80 ± 1.00        & 71.60 ± 0.70 & 37.50 ± 1.20     & 62.00 ± 1.30          \\
TransductCNAPS \cite{bateni2022enhancing} & 45.80 ± 1.00        & 72.50 ± 0.70 & 37.70 ± 1.10     & 62.90 ± 1.30          \\
Ours                      & \textbf{52.47 ± 1.03} & \textbf{75.92 ± 0.66}       & \textbf{42.36 ± 1.14} & 59.23  ±  1.27 \\
\hline
\end{tabular}
}
\end{table}

\begin{figure}[!t]
    \centering
    \includegraphics[width=0.96\linewidth]{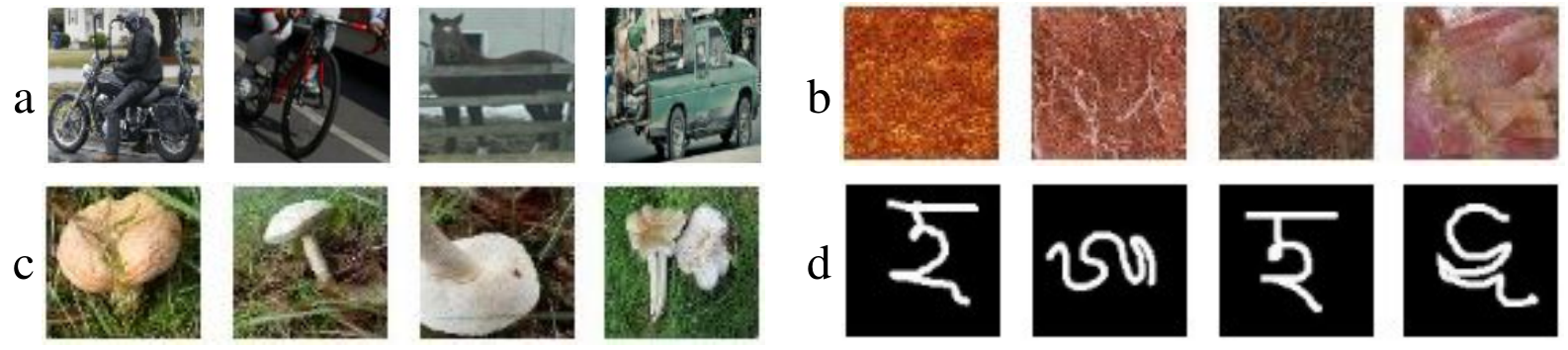}
    \caption{Image examples: (a) MSCOCO, (b) Textures, (c) Fungi, (d) Omniglot.}
    \label{fig:MetaExamples}
\end{figure}

We further evaluate the cross-domain performance of our model on a larger-scale dataset, Meta-dataset \cite{triantafillou2019meta}. Meta-dataset includes 10 different datasets. Different from the traditional N-way K-shot task, it does not fix the number of training samples for each class and the number of classes when sampling few-shot tasks. Instead, it randomly selects $N$ classes within a range $[5,50]$ and randomly selects a different number of training samples for each class within $[1,100]$. Here, we follow a cross-domain setting of Meta-dataset, i.e., pre-training on ImageNet ILSVRC 2012 \cite{2009ImageNet} and testing on  other datasets. 
Table \ref{tab:metadataset} reports the cross-domain evaluation results on four representative datasets, i.e., MSCOCO \cite{LinMBHPRDZ14COCO}, Describable Textures \cite{CimpoiMKMV14cvpr}, Fungi \cite{Schroeder18Fungi}, Omniglot \cite{Lake2015Science}. Some image examples of these datasets are also exhibited in Figure \ref{fig:MetaExamples}.
The MSCOCO dataset, like ILSVRC, is a semantically rich and diverse dataset, so  our method can better avoid domain bias when migrating base knowledge, thereby achieving good cross-domain performance on MSCOCO. 
It can also be observed from the results that our method achieves good cross-domain performance on the two fine-grained image datasets Textures and Fungi, verifying the effectiveness of our method. 
However, our method's cross-domain performance on the Omniglot dataset is relatively unsatisfactory compared to existing leading methods. This could plausibly be attributed to the circumstance that Omniglot is a multilingual alphabetic dataset where images solely comprise strokes (as depicted in Figure \ref{fig:MetaExamples}), thereby hindering our method's ability to proficiently exploit semantic information.

\begin{table}[!t]
\centering
\caption{Module analysis of our proposed framework on \textit{mini}ImageNet.}
\begin{tabular}{c|cc}
\toprule
Method  & 5-way 1-shot                       & 5-way 5-shot                       \\ \midrule
Baseline           & 64.71 ± 0.61                   & 78.34 ± 0.47                   \\ \hline
IVDH-g                & 65.23 ± 0.60                   & 79.01 ± 0.48                    \\
IVDH-x                & 65.36 ± 0.61                   & 81.53 ± 0.42                   \\
IVDH                & 65.92 ± 0.61                   & 82.45 ± 0.42                   \\ 
\hline
PVDH-p        & 67.15 ± 0.61                   & 83.91 ± 0.42                   \\
PVDH-v        & 68.61 ± 0.62                   & 84.12 ± 0.42                   \\
PVDH        & 70.17 ± 0.60                   & 84.53 ± 0.42                   \\ \hline
Ours-l               & 70.63 ± 0.60 &  84.20 ± 0.54 \\
Ours-full               & \textbf{70.64 ± 0.60} & \textbf{84.71 ± 0.41} \\
\hline
\end{tabular}
\label{tab:ablation}
\end{table}

\subsection{Ablation Study}
\label{sec:abs}

In this section, we conduct detailed experiments to analyze each module of our framework and the important factors that affect the performance.

\textbf{Module analysis:} 
Our proposed framework consists of two main modules, i.e., IVDH and PVDH. The baseline model of our method for classification is a WRN-28-10 backbone network with a classifier head, as described in Section \ref{sec:settings}. In ablation study, the baseline follows the same training procedure of our method except that our method generates hallucinated data for model training, since our method is dedicated to complementing the few-shot data appropriately. 
Table \ref{tab:ablation} shows the evaluation results of different modules of our framework on the \textit{mini}ImageNet dataset. IVDH represents the model performance by adding our instance-view data hallucination module onto the baseline, and it is analogous for PVDH.
As can be observed, the IVDH and PVDH modules improve the classification accuracy of the baseline by 1.21\% and 5.46\% on the 5-way 1-shot setting, respectively. As for the 5-way 5-shot setting, the performance gain is 4.11\% and 6.19\%. 
By integrating the IVDH and PVDH modules, our proposed method (Ours-full) further boosts the model performance to 70.64\% and 84.71\% accuracy for  1-shot and 5-shot settings, yielding  5.93\% and 6.37\% performance gain from the baseline, respectively.

We also conduct an ablation experiment on the effect of the semantic text $S_c$. While Ours-full uses the combination of the label and its definition in WordNet, let Ours-l denote a variant of our method that only uses the label for $S_c$. As observed from Table \ref{tab:ablation}, the two perform comparably, with Ours-full showing slightly better performance by using more semantic information.

In Table \ref{tab:ablation}, IVDH-g represents only using global semantic feature fusion to generate hallucinated samples in the IVDH module.
When only adopting global semantic feature fusion in IVDH, it shows that IVDH-g improves the baseline by 0.52\% and 0.67\% for the 1-shot and 5-shot scenarios, respectively. It also demonstrates local semantic correlated attention plays a more important role in IVDH. 
In \cite{wang2021mtunet}, MTUNet provides a visual explainable attention mechanism for few-shot learning without semantics. 
We implement their method in our IVDH module for comparison, denoted as IVDH-x in Table \ref{tab:ablation}. 
As can be observed, IVDH-x exhibits an improvement over IVDH-g but obtains a smaller performance gain than our IVDH, verifying the efficacy of our local semantic correlated attention mechanism.

PVDH-p denotes our PVDH module taking prototypes as hallucinated samples without data resampling. PVDH-v denotes our PVDH module without semantic modeling.
When using class prototypes as hallucinated samples without data resampling in PVDH, it shows that PVDH-p still performs better than IVDH, demonstrating the estimated prototypes serve as more stable samples. But it is worse than PVDH using data resampling, showing the model does benefit from the diversity of sampled data.
When not incorporating  semantic knowledge to guide the estimation of class prototypes and the associated distributions, PVDH-v exhibits a performance drop of 1.56\% and 0.41\% for the 1-shot/5-shot scenarios, respectively. It reflects the importance of using semantic relation guidance in PVDH.

\begin{table}
\centering
\caption{Effect analysis of our method on coarse-grained and fine-grained datasets (\textit{mini}ImageNet and CUB).} \label{tab:diffDataset}
\resizebox{1.01\linewidth}{!}{
\begin{tabular}{c|cc|cc}
\toprule
\multirow{2}{*}{Method} & \multicolumn{2}{c|}{\textit{mini}ImageNet} & \multicolumn{2}{c}{CUB}  \\
        & 1-shot       & 5-shot   & 1-shot     & 5-shot      \\
\midrule
Baseline &	64.71 ± 0.61  & 78.34 ± 0.47    &  80.63 ± 0.79 & 90.82 ± 0.43 \\
Ours &	\textbf{70.64 ± 0.60} & \textbf{84.71 ± 0.41} &	\textbf{81.57 ± 0.61} &	\textbf{91.30 ± 0.32} \\ 
\bottomrule
\end{tabular}
}
\end{table}

\begin{table}[!t]
\centering
\caption{Comparison results of our method using different backbones on \textit{mini}ImageNet.} \label{tab:backbone_acc}
\resizebox{1.01\linewidth}{!}{
\begin{tabular}{c|cc|cc}
\toprule
\multirow{2}{*}{Backbone} & \multicolumn{2}{c|}{Baseline} & \multicolumn{2}{c}{Ours} \\
                          & 1-shot        & 5-shot        & 1-shot      & 5-shot      \\ \midrule
Conv4                     & 39.21 ± 0.55    & 59.06 ± 0.52    & 49.48 ± 0.56  & 65.79 ± 0.50  \\
Conv6                     & 47.00 ± 0.56    & 63.25 ± 0.55    & 52.30 ± 0.59  & 68.66 ± 0.53  \\
ResNet10                  & 53.77 ± 0.63    & 71.82 ± 0.53    & 62.03 ± 0.62  & 77.36 ± 0.48  \\
ResNet18                  & 53.08 ± 0.61    & 71.87 ± 0.50    & 62.19 ± 0.62  & 77.79 ± 0.48  \\
ResNet34                  & 53.25 ± 0.62    & 71.75 ± 0.51    & 61.89 ± 0.61  & 78.00 ± 0.48  \\
WRN-28-10              & \textbf{64.71 ± 0.61}    & \textbf{78.34 ± 0.47}    & \textbf{70.64 ± 0.60}  & \textbf{84.71 ± 0.41}  \\
\hline
\end{tabular}
}
\end{table}

We further conduct an experiment to evaluate the effectiveness of our method on different kinds of datasets. Specifically, we compare on the coarse-grained \textit{mini}ImageNet dataset and the fine-grained CUB dataset, and the results are reported in Table \ref{tab:diffDataset}. As can be observed, the proposed method shows obvious effect on the coarse-grained datasets like \textit{mini}ImageNet. In the case of the fine-grained CUB, where all categories are birds and the difference in their semantic space is slight, our method shows less obvious effect. However, it is worth noting that the difference in the visual space is also modest in such cases. Thus the semantic information we explore can also take some effect, which is corroborated by the observed performance gain on CUB. 

\textbf{Performance with different backbones:} 
We use different backbone networks as feature extractors in our method to evaluate their influences on the model performance. As shown in Table \ref{tab:backbone_acc}, six different backbones are adopted for evaluation on the \textit{mini}ImageNet. For comparison, the results of the baseline are also reported. It can be easily observed that our method can consistently outperform the baseline by a large margin (about 5\%-10\%) for 1-shot/5shot settings, soundly verifying the effectiveness of our proposed framework. Furthermore, it can also be inferred that, for simple backbone networks (conv4, conv6), the increase of network depth can effectively improve the performance of the model. However, with respect to deeper networks like ResNet, simply increasing the depth is not helpful to improve the accuracy. For example, ResNet34 is much deeper than ResNet10, but its performance is witnessed a slight  drop. One important reason is lacking efficient training data.
In contrast, WideResNet is observed to be more suitable for few-shot learning tasks, by increasing the width of the network and applying dropout to avoid overfitting. As revealed in Table \ref{tab:backbone_acc}, WRN-28-10  yields obvious performance gain than ResNet34.

\begin{table}[!t]
\centering
\caption{5-shot accuracy of different merging strategies in PVDH on \textit{mini}ImageNet.}\label{tab:mergeStrategy}
\begin{tabular}{l|l}
\toprule
Method                                       & Accuracy                \\
\midrule
w/o merging                   & 84.39 ± 0.41                   \\
merging before estimation & 84.57 ± 0.41                   \\
merging after estimation & \textbf{84.71 ± 0.41} \\
\hline
\end{tabular}
\end{table}

\textbf{Merging strategies for multiple shots in PVDH:}
In the PVDH module, we merge the candidate prototypes obtained from multi-shot samples  to yield the final prototype of a novel class, denoted as ``merging after estimation''. In fact, different merging strategies can be used. One strategy is not merging the multiple candidate prototypes, which results in representing the data distribution as a mixture of Gaussians for data resampling, denoted by ``w/o merging''. Another strategy is merging the multiple sample features into one by averaging and then estimating the prototype based on the merged feature, denoted by ``merging before estimation''. 
We conduct experiments to evaluate these merging strategies.
As shown in Table \ref{tab:mergeStrategy}, our method using the three strategies exhibits very close performance, and the ``merging after estimation'' strategy we adopt offers the best performance.

\textbf{Effect of important parameters:}
In this part, we conduct experiments to analyze how the important parameters affect the model performance. We evaluate a certain parameter while keeping the other parameters fixed.

\begin{figure}[!t]
    \centering
    \includegraphics[width=0.8\linewidth]{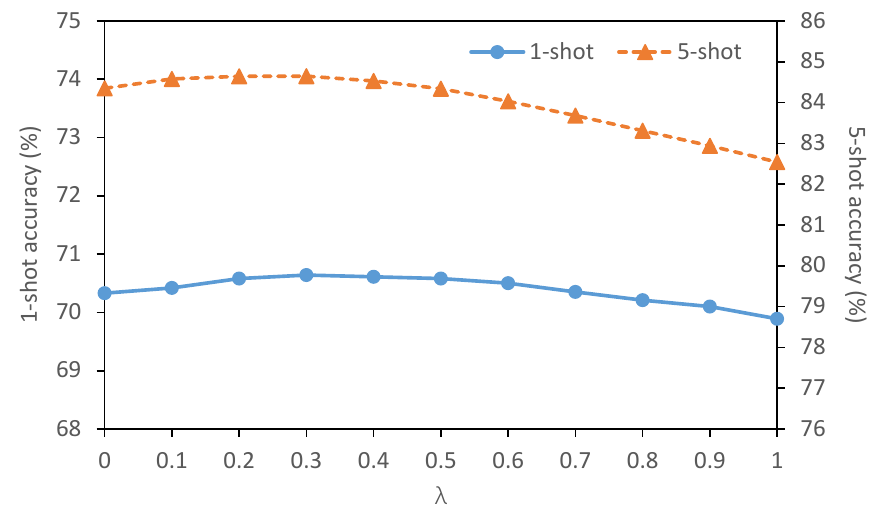} 
    \includegraphics[width=0.8\linewidth]{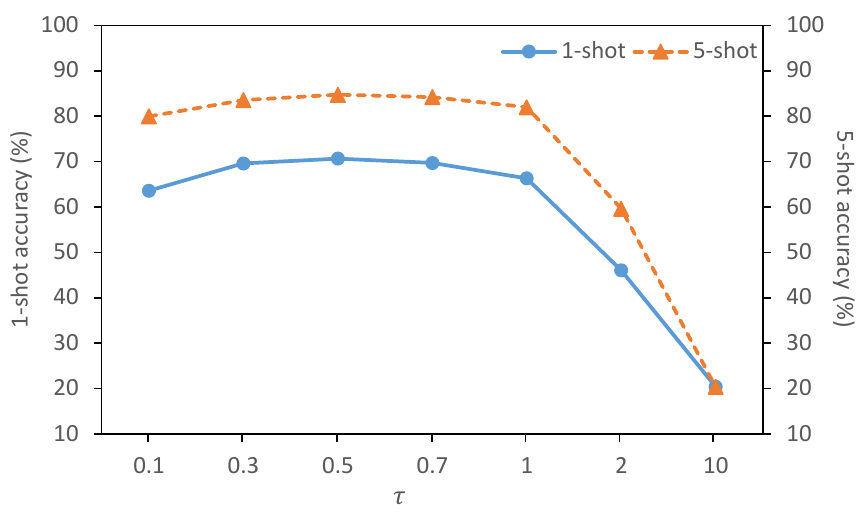}
    \caption{Effect analysis of $\lambda$ and $\tau$ on \textit{mini}ImageNet.}
    \label{fig:paramFigLambda}
\end{figure}

As shown in Figure \ref{fig:paramFigLambda}, we conduct an experiment on the \textit{mini}ImageNet dataset to evaluate the effect of the parameter $\lambda$ that controls the strength of semantic feature fusion in IVDH. It can be observed that the model performance in both the 1-shot and 5-shot scenarios first increases and then decreases gradually when challenging the value of $\lambda$ from 0 to 1, and the best performance is observed when $\lambda$ is around 0.3. Therefore, we set $\lambda$ to 0.3 in our experiments.
As for the parameter $\tau$ for Turkey’s Ladder of Powers transformation, we follow previous work to set it as 0.5. We conducted an experiment to verify its influence on our model. As shown in Figure \ref{fig:paramFigLambda}, the model attained the highest accuracy in the 1-shot and 5-shot settings when $\tau$ was approximately 0.5.

\begin{figure*}[!t]
    \centering
    \includegraphics[width=0.99\linewidth]{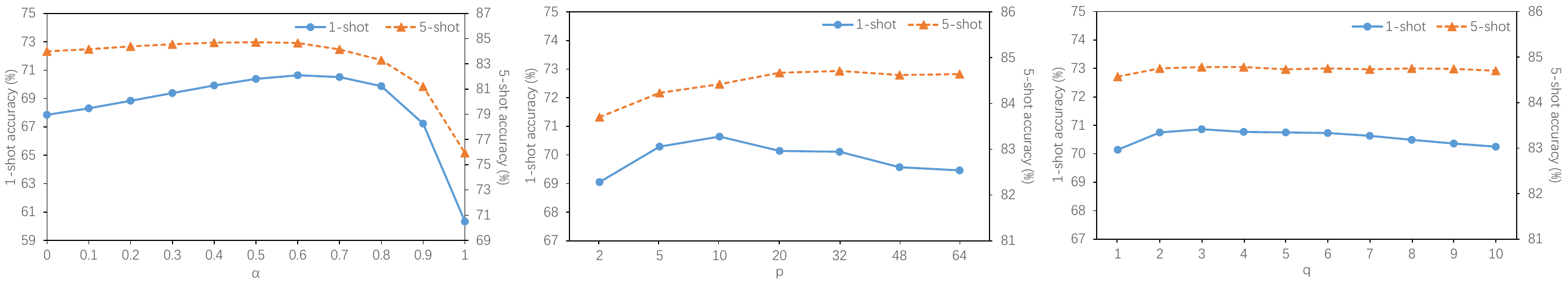}
    \caption{Effect analysis of the parameters $\alpha$, $p$, and $q$ in PVDH  on \textit{mini}ImageNet.}
    \label{fig:paramFigPVDH}
\end{figure*}

Figure \ref{fig:paramFigPVDH} reports the evaluation results on \textit{mini}ImageNet for analyzing the important parameters $\alpha$, $p$, and $q$ in the PVDH module. With respect to the parameter $\alpha$ that controls the fusion degree in prototype estimation, when changing the value from 0 to 1, it first improves the model accuracy gradually, reaching a peak at around 0.6. Afterwards, obvious performance drop is witnessed when further increasing $\alpha$ to 1.
Some conclusions can be drawn from the observations.
On the one hand, it shows the important benefits of incorporating similar basses to estimate the prototypes of novel classes. And the performance gain is more obvious when the available samples are fewer, as revealed by the performance comparison of 1-shot and 5-shot settings. On the other hand, it also reflects that severe noise would be introduced to make a negative impact on the model if excessively relying on the information of base classes.

As for the parameters $p$ and $q$, which involve in defining the semantic-visual similarity metric for selecting the most similar base classes, different phenomenons are observed. The two parameters reflect the number of selected base classes and should take a positive integer value. Take a look at $p$ first. It determines the range to select base classes semantically. 
As exhibited in Figure \ref{fig:paramFigPVDH}, it gradually increases the model performance when $p$ takes a larger value. the model performance appears to reach a plateau when $p$ is around 32. When further increasing $p$, the performance exhibits a slight fluctuation but is relatively steady. However, it shows a quite different trend in the 1-shot scenario. As can be observed, when $p$ gradually takes a larger value, the model performance has an obvious climb and reaches a peak when $p$ is around 10. Then the performance becomes to drop gradually when $p$ further increases. It can be inferred that the semantic selection is more critical when the samples are fewer, where a relatively small $p$ can help better filter out the interference caused by the distracting factors contained in the few image samples.
With respect to the parameter $q$ for selecting base classes visually, its influence on the model performance is relatively small when taking different values. As shown in Figure \ref{fig:paramFigPVDH}, when $q$ takes a value from 1 to 10, the model performance in the 5-shot scenario first has a slight increase and then plateaus with a very small fluctuation. While in the 1-shot scenario, the model performance reaches a peak when $q$ is around 2 or 3, and shows a slightly decreasing trend when $q$ becomes larger. Based on these observations, setting $q$ to 2 or 3 is a reasonable choice.

\begin{figure}[!t]
    \centering
    \includegraphics[width=0.85\linewidth]{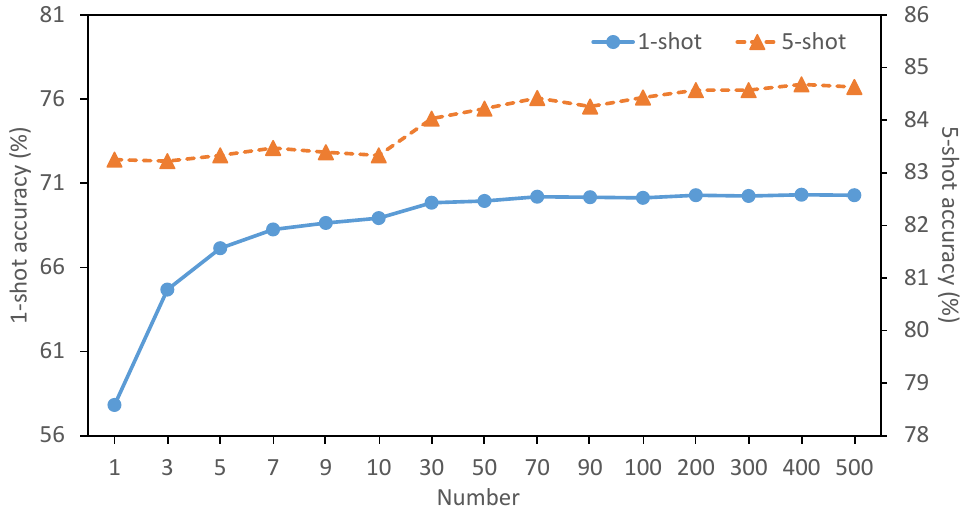}
    \caption{Effect analysis of the number of resampled data  on \textit{mini}ImageNet.}
    \label{fig:paramFigSampling}
\end{figure}

One important characteristics of the PVDH module is that it can resample data from the distribution associated with the estimated prototype of a novel class and thus generate a large number of hallucinated data. We conduct experiments using only the PVDH module to analyze how the number of resampled data affects the model performance.
As shown in Figure \ref{fig:paramFigSampling}, it is consistent with our intuition that the model performance continually improves as the number increases. Nevertheless, different increasing trends are observed for the 1-shot and 5-shot scenarios. In the 1-shot scenario, the performance increase is dramatically sharp when the number is not larger than 5. It should also be noticed that serious negative impact on the model could happen when the number is 1, since it may bring intractable noise for the 1-shot sample. As the number increases to around 30, the model performance appears to plateau, and increasing the number to a much larger value such as 500, slight improvement is observed but is negligible. The reason may be that the distribution estimated from the 1-shot sample of a novel class is less informative and more resampled data from the distribution cannot offer more useful information of the corresponding novel class. 
As for the 5-shot scenario, the performance improvement is not obvious when the number is relatively small (e.g., less than 10). That may be because there already exist 5-shot samples for each class. 
But when the number takes a larger value like 30, an impressive performance gain is witnessed. The performance appears to be quite steady when the number reaches about 200, and a larger number brings little improvement.

\subsection{Visualization Analysis}
In this section, visualization results are presented to further analyze each module of our proposed framework qualitatively.

\begin{figure}[!t]
    \centering
    \includegraphics[width=0.92\linewidth]{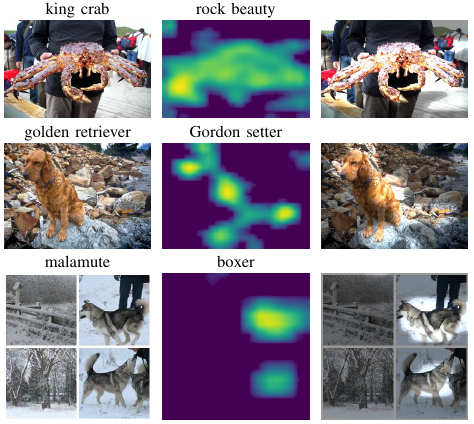}
    \caption{Visualization of several image samples hallucinated by local semantic correlation attention in IVDH. Each row from left to right shows the original image sample of a novel class, the attention map activated from its most semantically correlated base class (denoted by the class label above it), and the hallucinated image sample.}
    \label{fig:SCAvis}
\end{figure}

\textbf{Visualization from IVDH:}
As described in Section \ref{sec:IVDH}, the IVDH module hallucinates image samples of novel classes by exploiting the spatial attention maps activated from the highly semantically correlated base classes.
Figure \ref{fig:SCAvis} visualizes several hallucinated image samples generated by the semantic correlation attention for qualitative analysis. 
It should be noted that the hallucinated samples are visualized after some processing, since the original image is rescaled to a fixed size before being input to the model and leads to an attention map with a size different from the original size. The class labels in the second column are the most semantically correlated base classes of the corresponding novel classes in the first column. It can be observed that a plausible attention map is likely to be generated by the semantically selected base class and results in a hallucinated image sample that emphasizes some important regions for recognizing the corresponding class. 

\begin{figure}[!t]
    \centering
    \includegraphics[width=0.98\linewidth]{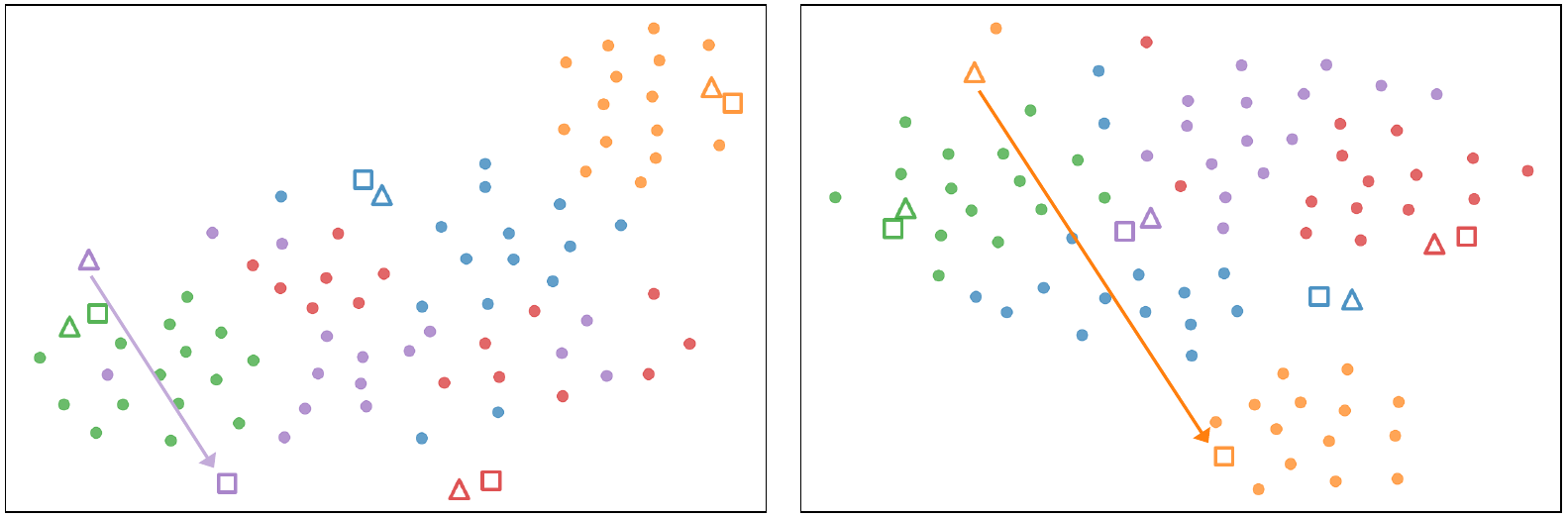}
    \caption{t-SNE visualization for global semantic feature fusion of IVDH in handling 5-way 1-shot tasks. Different colors denote different classes. Triangles denote the 1-shot samples (support set) of novel classes. Squares denote the hallucinated samples by global semantic feature fusion. Circles denote the query samples (query set). Best viewed in color.}
    \label{fig:SFIvis}
\end{figure}

The IVDH module uses global semantic feature fusion to hallucinate samples by projecting them closer to their class prototypes in the feature space via semantic embedding fusion. 
Figure \ref{fig:SFIvis} illustrates two t-SNE visualization examples of this mechanism in handling 5-way 1-shot tasks on \textit{mini}ImageNet.
Samples from different classes are denoted by different colors in the figure. Take a close look at the original 1-shot sample (denoted by triangles), the hallucinated sample (denoted by squares) and the query samples (denoted by circles) of each novel class, we can observe that the hallucinated sample feature generated by semantic feature fusion commonly lies next to the original sample feature for general cases, especially when the sample is within the potential class cluster. 
It can be found that the biggest advantage of this mechanism is dealing with the abnormal samples. As exhibited in Figure \ref{fig:SFIvis}, for some samples that severely deviate from the distribution and lie more near the clusters of some other classes, it is likely to obtain a hallucinated sample more near the cluster of its class rather than those of other classes (denoted by the arrows), thus providing samples with higher confidence for the model to learn a better classification boundary.

\begin{figure}
    \centering
    \includegraphics[width=0.99\linewidth]{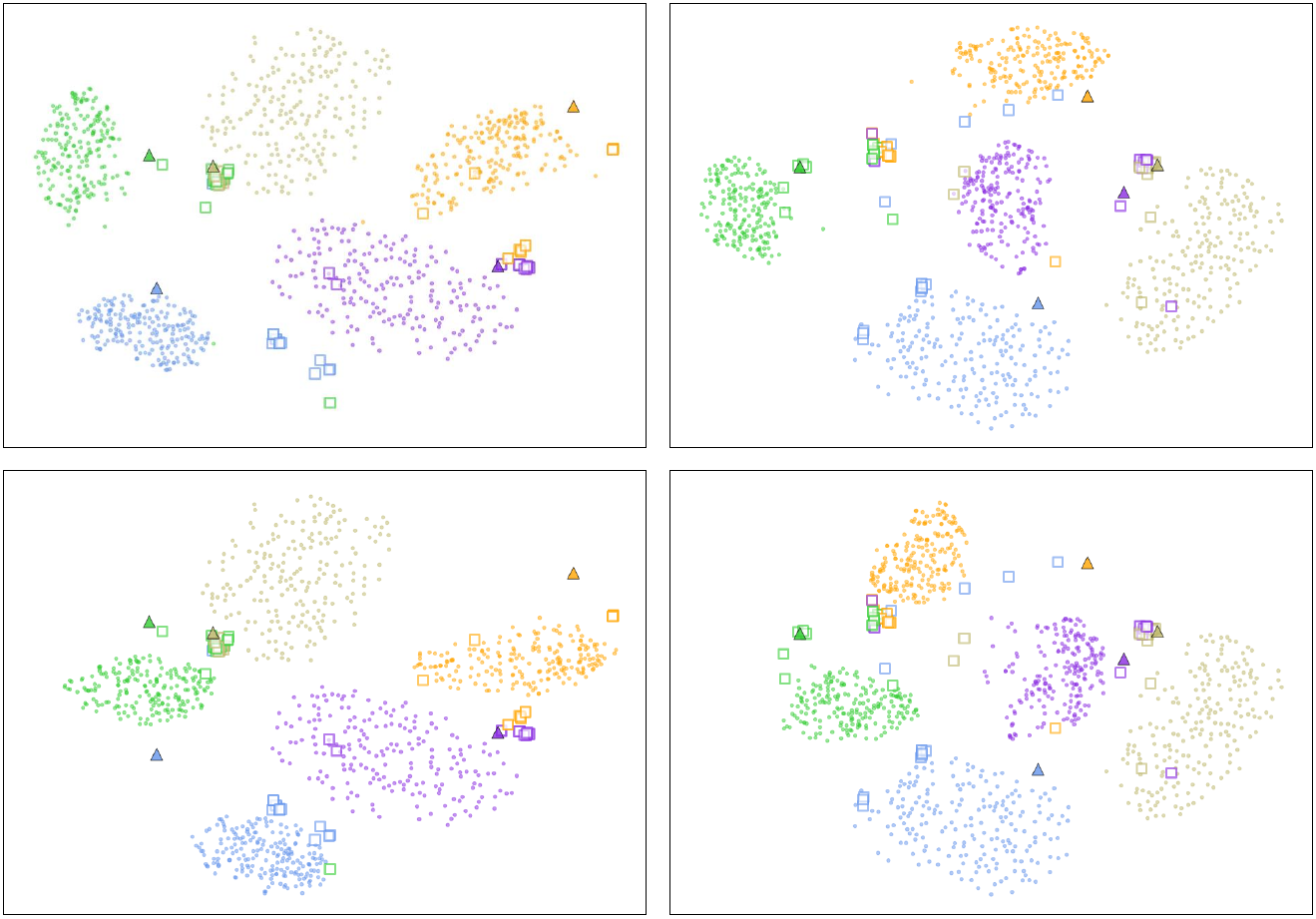}
    \caption{t-SNE visualization of resampled data by PVDH in handling 5-way 1-shot tasks. The upper and bottom rows show the results generated by the PVDH module without and with the semantic similarity metric, respectively. Different classes are denoted by different colors. Triangles denote the training samples (support set) of novel classes. Squares denote the query samples (query set). Circles denote the resampled data generated by PVDH. Best viewed in color.}
    \label{fig:PVDHvis}
\end{figure}

\textbf{Visualization from PVDH:} 
The PVDH module is able to generate a large number of feature samples to compensate for the data deficiency, which brings an important impact on improving the model. 
Figure \ref{fig:PVDHvis} shows several t-SNE visualization examples of the resampled data generated by the PVDH module in handling 5-way 1-shot tasks on \textit{mini}ImageNet.
To better analyze the effect of the semantic metric in PVDH, we show the visualization results of our PVDH module without and with semantic metric for comparison. 
As can be observed, by using the prototypes of highly similar base classes to help estimate the prototypes of novel classes, the data resampled from the associated distributions can roughly distribute near the query samples of the same classes for common cases. 
The visualization results also demonstrate that using the semantic  metric can help better estimate the class prototypes, especially when the available few samples severely deviate from the distribution, as illustrated by the blue cluster in the first column and the yellow cluster in the third column.

\subsection{Discussion and Potential Extensions}
\label{sec:Limitation}

In the previous subsections, extensive experimental comparisons and analyses have shown the effectiveness of our method. However, certain limitations may arise in some cases, as we observed with the Omniglot dataset (illustrated in Figure \ref{fig:MetaExamples}), where images and labels offer limited semantic information, leading to less satisfactory performance of our method. In future work, we will investigate potential solutions to overcome these limitations and improve the generalizability of our method.
Additionally, there are several other aspects of our method that merit further exploration. Firstly, in the prototype-view data hallucination module, we compute the semantic and visual similarity in their respective domains and combine them via ranking to model the semantic-visual relationship. It is worth exploring better semantic-visual relation modeling schemes, such as directly bridging the two domains as in CLIP \cite{RadfordKHRGASAM21icml}.
Secondly, we plan to investigate the scalability of our proposed method beyond images and explore its potential for few-shot video classification \cite{HuGX21tmm}. Actually, our method can be directly applied to this task by replacing image features with video features. However, the complex spatio-temporal context of videos may pose challenges, and developing proper mechanisms to better exploit such context may be needed for effective hallucination.
While Gao et al. \cite{GaoZX21pami} introduce an innovative method to learn prototypes from a group of semantic word vectors for zero-shot video classification, we plan to incorporate visual information to build more effective prototypes of video categories for data hallucination in few-shot scenarios, thereby better exploiting the advantages of our method.

\section{Conclusion}\label{sec:Conclusion}

In this work, we propose a novel framework that exploits semantic relations to guide dual-view data hallucination for few-shot image recognition.
Specifically, the proposed framework transfers the information of base classes to generate diverse and reasonable data samples of novel classes from both instance and prototype views to promote model learning. 
On the one hand, an instance-view data hallucination module generates hallucinated samples from each instance by employing local semantic correlated attention and global semantic feature fusion. 
On the other hand, a prototype-view data hallucination module exploits semantic-aware measure to estimate the prototypes of novel classes and the associated distributions, which thereby harvests prototypes as more stable samples and enables resampling a sufficient number of samples. 
Extensive experiments and comparisons conducted on several popular few-shot datasets demonstrate that the proposed framework is quite effective and outperforms state-of-the-art methods.
We believe that our work could shed some new light on the research of few-shot learning. The data-oriented paradigm of our work makes it  feasible to combine with other kinds of FSL methods to yield better models.

\bibliographystyle{IEEEtran}
\bibliography{reference}

\end{document}